\documentclass[letterpaper, 10 pt, conference]{ieeeconf}  

\IEEEoverridecommandlockouts                              %
\usepackage[colorlinks,linkcolor=green,citecolor=red,urlcolor=blue,bookmarks=true,hypertexnames=true]{hyperref} 
\usepackage[utf8]{inputenc}

\usepackage{amsthm}
\usepackage[T1]{fontenc}
\usepackage{soul, color, xcolor}
\soulregister{\cite}7 
\soulregister{\citep}7 
\soulregister{\citet}7 
\soulregister{\ref}7 
\soulregister{\pageref}7 
\usepackage{graphicx} 
\usepackage{epsfig} 
\usepackage{mathptmx} 

\usepackage{amsthm,amssymb}
\usepackage{mathrsfs}
\usepackage{verbatim}
\usepackage{amsmath,amsfonts}
\usepackage{algorithmic}
\usepackage{algorithm}
\usepackage{booktabs}
\usepackage{url}
\usepackage{multirow}
\usepackage{multicol}
\usepackage{balance}

\title{\LARGE \bf
Semantic SLAM with Rolling-Shutter Cameras and Low-Precision INS in Outdoor Environments}

\author{Yuchen Zhang$^{1,\dagger}$, Miao Fan$^{2, \dagger,*}$, Shengtong Xu$^{3}$, Xiangzeng Liu$^{4}$, and Haoyi Xiong$^{5}$  
\thanks{$^{1}$Engineer at NavInfo Co. Ltd., China.}
\thanks{$^{2}$Chief scientist at NavInfo Co. Ltd., China. Senior member of IEEE.}
\thanks{$^{3}$Principal product manager at Autohome Inc., China.}
\thanks{$^{5}$Associate professor at Xidian University, China.}
\thanks{$^{5}$Principal scientist at Baidu Inc., China. Senior member of IEEE.}
\thanks{$\dagger$Equal contribution.}
\thanks{*Correspondence: {miao.fan@ieee.org}.}
}
        
\begin{document}

\maketitle
\thispagestyle{empty}
\pagestyle{empty}

\begin{abstract}
Accurate localization and mapping in outdoor environments remains challenging when using consumer-grade hardware, particularly with rolling-shutter cameras and low-precision inertial navigation systems (INS). We present a novel semantic SLAM approach that leverages road elements such as lane boundaries, traffic signs, and road markings to enhance localization accuracy. Our system integrates real-time semantic feature detection with a graph optimization framework, effectively handling both rolling-shutter effects and INS drift. Using a practical hardware setup which consists of a rolling-shutter camera (3840×2160@30fps), IMU (100Hz), and wheel encoder (50Hz), we demonstrate significant improvements over existing methods. Compared to state-of-the-art approaches, our method achieves higher recall (up to 5.35\%) and precision (up to 2.79\%) in semantic element detection, while maintaining mean relative error (MRE) within 10cm and mean absolute error (MAE) around 1m. Extensive experiments in diverse urban environments demonstrate the robust performance of our system under varying lighting conditions and complex traffic scenarios, making it particularly suitable for autonomous driving applications. The proposed approach provides a practical solution for high-precision localization using affordable hardware, bridging the gap between consumer-grade sensors and production-level performance requirements.

\end{abstract}
{\keywords semantic SLAM, visual odometry, rolling-shutter cameras, autonomous driving.}

\section{Introduction}
Visual Simultaneous Localization and Mapping (VSLAM) represents a fundamental challenge in computer vision and robotics, with critical applications ranging from autonomous navigation to self-driving vehicles \cite{r1,r2,r3}. VSLAM systems aim to simultaneously construct a map~\cite{r0,r00} of an unknown environment while tracking camera position and orientation using visual data. However, these systems often struggle with cumulative errors, particularly in challenging conditions such as unstable lighting or dynamic scenes. 

\begin{figure*}[htp!]
  \includegraphics[width=\textwidth]{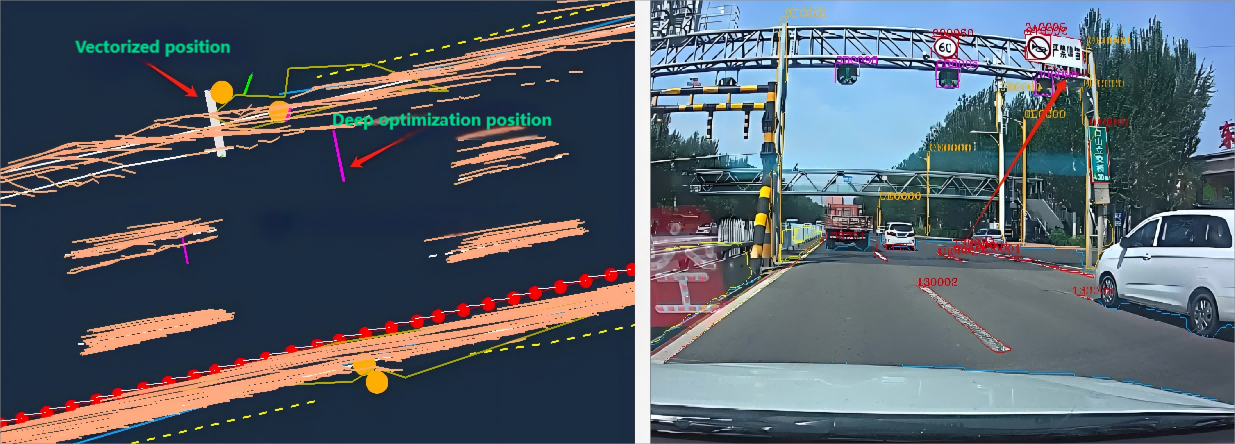}
  \vspace{-5mm}
  \caption{A case in point in a highway scenario. Camera position accuracy has a great impact on accuracy, which is huge on ordinary roads and high speeds. IMU accuracy is extremely poor, after the GNSS signal is lost for 3 seconds, the combined navigation result is dispersed, depth optimization is not effective, and the relative position error is large after vectorization of traffic signs in the same lateral position.}
  \label{fig:1}
\end{figure*}

To address these limitations, Visual-Inertial Navigation Systems (VINS) integrate data from visual sensors with high-frequency measurements from Inertial Measurement Units (IMU). This fusion compensates for temporary losses in visual tracking and enhances system stability. However, VINS faces significant challenges in outdoor environments where the accuracy of both ground and airborne feature detection impacts system performance \cite{r4}. Camera pose estimation errors can propagate through the system, affecting object localization and overall map reconstruction quality. Moreover, inaccurate camera calibration can compromise sensor fusion, further degrading system performance. These challenges become particularly acute in resource-constrained devices. Limited computational power prevents effective processing of high-frequency sensor data, impacting system responsiveness and accuracy. 
As illustrated in Fig. \ref{fig:1}, when using lower-precision IMUs, motion estimation errors accumulate over time, leading to significant drift in positioning results. Furthermore, outdoor environments present additional challenges due to unreliable Global Navigation Satellite System (GNSS) signals. Signal occlusion or interference prevents GNSS-based position corrections, potentially resulting in severe navigation errors as positional drift accumulates unchecked.

Mainstream SLAM methods typically require high-performance hardware for efficient feature extraction, matching, and optimization. For instance, ORB-SLAM \cite{r5} estimates camera poses by comparing matched key points from the camera input with recent keyframes. The algorithm first extracts key points from the camera input, which are selected image features based on predefined filters \cite{r5, r6} or machine learning models \cite{r7, r8}, and tracks them across keyframes by matching descriptors. Pose estimation is then performed by minimizing the reprojection error between the matched 3D map points in world coordinates and the 2D key points. To ensure reliable pose estimation, many key points must be matched across consecutive frames, making scene continuity in neighboring frames crucial. Camera images must be captured and processed frequently to maintain high similarity between successive frames. However, frequent image processing and keypoint comparisons lead to significant computational overhead, while recording all keyframes increases storage complexity. As a result, low-computing-power devices often struggle to meet these computational demands, leading to inefficient and unresponsive systems that may fail to function properly under resource constraints.

The ORB-SLAM algorithm \cite{r9} has several limitations in practical applications. First, the ORB feature descriptor is a binary descriptor based on luminance contrast, making it difficult to capture high-level or semantic content in images. Second, the optimization process in ORB-SLAM primarily minimizes reprojection error, which relies solely on geometric relationships and lacks constraints from external factors, such as the physical environment and semantic information. As a result, there may be insufficient evidence to ensure the reliability of the global optimal solution. Additionally, perspective distortion can affect geometric feature matching, especially in scenarios with large angles or rapid motion, where feature point changes exceed the descriptor’s capacity, leading to inaccurate matching. In low-light conditions or adverse weather, image contrast and texture information decrease, significantly reducing the accuracy of feature point extraction and matching. Moreover, while distant feature points are generally easier to match, they tend to have lower resolution, resulting in poorer quality and accuracy. This leads to less reliable matching results, which contribute less to positional and attitude estimation and may even introduce errors.

To significantly reduce computational resources and costs while addressing the inherent limitations of ORB-SLAM algorithms, this study introduces a novel algorithm that leverages low-cost, low-power rolling-shutter cameras and low-precision INS. The algorithm utilizes semantic information to perform visual odometry and mapping in complex outdoor environments. Unlike traditional methods, it replaces feature points with outdoor road semantic elements, such as lane markings and traffic signs. These elements are first identified and tracked by an AI perception and tracking model, while INS trajectories provide coarse-grained, low-precision semantic vectorization. Residual errors are minimized by back-projecting the semantic elements into the camera coordinate system. The proposed method handles scenarios involving traffic signs, poles, lane markings, etc., and achieves high-precision localization and fine-grained mapping through graph optimization for diverse conditions. The key contributions of this paper are summarized as follows:
\begin{itemize}
\item The proposed method allows the use of extremely low-cost roll-up shutter cameras and low-precision INS, which significantly reduces the hardware cost and energy consumption compared to radar sensors or global shutter cameras, and provides new ideas and potential for large-scale applications of SLAM technology.

\item Introduced a new method to implement Visual Odometry by using road elements such as lane lines, traffic signs, crosswalks, and other alternative feature points, and completed road element mapping simultaneously.

\item Compared with the baseline method, the recall rate and precision rate of the proposed method are improved by up to 5.35\% and 2.79\%, respectively. The average relative position precision between semantic elements is kept within 10cm.
\end{itemize}

\section{Related Work}
Recent SLAM research has evolved from purely geometric approaches to systems incorporating semantic understanding and multi-sensor fusion. This section reviews key developments in semantic SLAM, focusing on outdoor navigation systems and practical implementations with consumer-grade hardware.
\subsection{Visual SLAM}
Earlier VSLAM methods used predefined descriptors such as SIFT \cite{r18} to extract feature points from the image \cite{r17}, and the collection of frames consisting of these feature points forms the basis of the map. The camera pose is estimated by minimizing the reprojection error between the query frame and the stored frame with the most matching feature points. However, to ensure that a large number of map points can be found, a large number of keyframes must be stored in the map and the observations must have significant continuity. As a result, these methods typically require significant computational and memory resources. 

ORB-SLAM enhances the performance of VSLAM techniques \cite{r5}, which introduces the ORB features and applies them to the tasks of parallel tracking, mapping, relocalization, and loop closing using optimizations based on bitmap optimization and bundle adjustment. These features perform well in short- and medium-term data correlation, enabling efficient visual SLAM. Subsequently, ORB-SLAM3 \cite{r9} introduces an IMU-based visual-inertial odometry (VIO) to support SLAM with monocular, binocular, and RGB-D cameras, which is capable of estimating camera trajectories and 3D structures in dynamic scenes through joint optimization, and long-time and wide-range scenes in which robustness is significantly improved.

In recent work, \cite{r31} proposed a framework of tightly coupled fusion of visual odometry and a vector high-definition map~\cite{r0} to solve the problem of unstable positioning results in the case of sparse observation and large noise. The algorithm used a sliding window to observe visual feature points and vector high-definition map landmarks and optimized their residuals in a tightly coupled manner. Promising experimental results are achieved in two challenging scenarios with noisy and sparse landmark observations. \cite{r32} proposed a monocular vision localization method using a vector map as the localization layer to detect semantic traffic elements from the image and match them with vectors in the map. To reduce the harmful problem of false matches, a non-explicit and differentiable data association process is implemented by aligning the vector mapping with the semantically detected distance transform, which enables the system to achieve centimeter and submeter accuracy in lateral and longitudinal directions, respectively.

\subsection{Semantic SLAM}
Semantic elements in the transportation environment can be used as additional information to enhance SLAM system performance, \cite{r24} utilizes a 3D semantic point cloud with landmark information to fuse the 3D map with relevant semantic information through coordinate system transformation and Bayesian updating. Fuzzy affiliation based on the Gaussian distribution is used to realize landmark data fusion, and a topological semantic map is constructed from it. \cite{r30} introduces a monocular visual positioning method that directly uses vector maps as the positioning layer by detecting semantic traffic elements from images and matching them with vectors in the map. This enables the system to achieve centimeter and submeter accuracy in horizontal and vertical directions, respectively. \cite{r25} presented an end-to-end visual-inertial odometry system that utilizes semantic features extracted from RGB-D sensors to create a semantic map of the environment, progressively refines the semantic map, and improves attitude estimation. \cite{r26} proposed a DS-SLAM system that combines a SegNet semantic segmentation network with a movement consistency check to filter out unstable parts of a dynamic scene, and realizes real-time robust semantic SLAM by running five threads in parallel for tracking, semantic segmentation, and local mapping. \cite{r27} proposed a YOLO-SLAM that uses lightweight YOLOv3 for target detection to provide semantic information in dynamic environments, and also removes feature points within the contours of dynamic objects using a deep transact screening method to minimize the impact of dynamic targets on the SLAM system.
\begin{figure}[h!]
  \centering
  \includegraphics[width=\columnwidth]{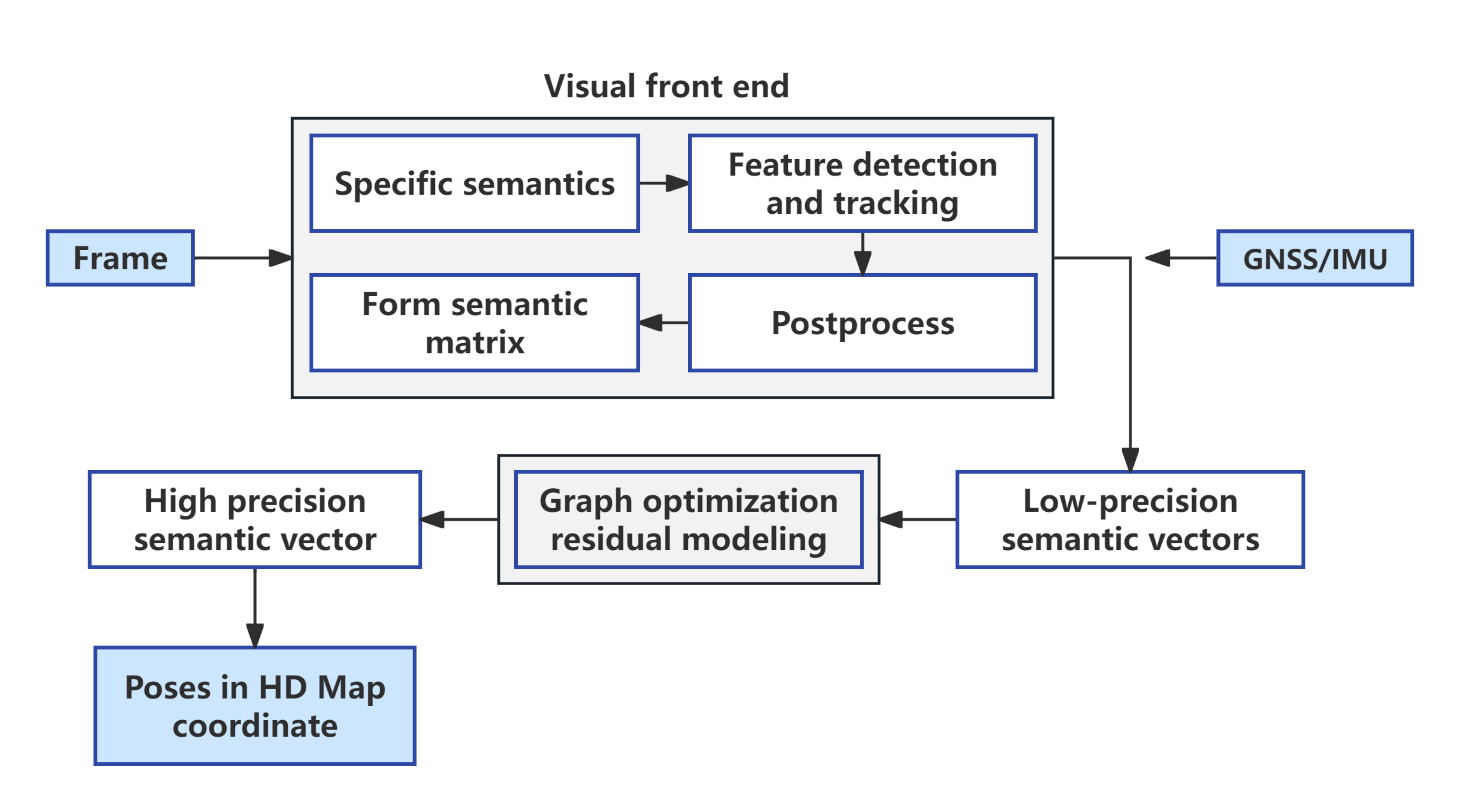}
  \caption{Schematic of semantic VO graph optimization.}
  \vspace{-4mm}
  \label{fig:VO}
\end{figure} 

Unlike the above methods, we use low-cost, low-power rolling shutter cameras and low-precision INS. At the same time, we do not rely on the matching of keyframes in visual SLAM or the comparison between the query frame and the reference frame, but instead, we replace the feature points with the semantic elements of the road outside the user. Semantic slam is scene-independent, meaning that it does not need to be retrained for different environments.

\section{Method}
\subsection{System Overview}
Our system integrates temporal, visual, and geographic information to construct a semantic map with precise geographic coordinates. The pipeline begins by correlating camera positions with vehicle ENU coordinates through timestamp matching, providing geographic context to each frame. Semantic elements are detected using YOLOv9 \cite{r28}, which extracts semantic information in the form: $Sem_j = (n_j ,x_j ,y_j ,w_j ,h_j )$, $j \in (1,2,....m)$, where $m$ denotes the number of semantic elements in the current frame, $n_j$ represents the semantic category, $(x_j, y_j)$ specify the top-left corner coordinates, and $(w_j, h_j)$ define the width and height of the semantic element's bounding box in pixel coordinates. A target tracking network \cite{r29} monitors these elements across consecutive frames, maintaining consistent identification and position information.

The system constructs a semantic element benchmark library by combining camera poses $\mathbf{T}^{c_i}_w$, geographic coordinates $\mathbf{E}_i$, and semantic information ${Sem_1, Sem_2, ..., Sem_m}$ from each frame. This library is the foundation for our semantic visual odometry system, as illustrated in Fig. \ref{fig:VO}.
\subsection{Semantic Feature Detection and Tracking}
In the localization stage, the real-time environment image information is obtained by the front camera of the vehicle, and the image information is input to the vision SLAM system to obtain the current camera position $\mathbf{T}_w^{c_x}$, through YOLOv9 network to obtain pixel semantic information, $\begin{Bmatrix}
    Sem'_1, Sem'_2,..., Sem'_w
\end{Bmatrix} $, $Sem'_o = (n'_o ,x '_o ,y'_o ,w'_o ,h'_o ) ,o \in (1 ,2,3 ,...,w)$. The two translation matrices  $\mathbf{t}_w^{c_x}=(x_x,y_x,z_x)^\mathrm{T}$ and $\mathbf{t}_s^{i}=(x_i,y_i,z_i)^\mathrm{T}$ are obtained separately by $\mathrm{\mathbf{T}}_w^{c_x}$ and semantic element frame position $\mathbf{T}_s^{i}$, $ i \in(1,2,..., n)$ respectively. Semantic element frame matching is done by violently retrieving semantic element frames from the benchmark library that satisfy the following public notices as candidate semantic element matching frames $\begin{Bmatrix}
    l_{a_1} ,l_{a_2} ,...,l_{a_k}
\end{Bmatrix} $:
\begin{equation}
    \delta  >\sqrt{(x_x-x_i)^2+(y_x-y_i)^2+(z_x-z_i)^2}
\end{equation}
where $\delta $ is the set distance threshold. Let $l_{a_i}, i \in (1, 2, ... , k)$ frames corresponding to the semantic information as $\begin{Bmatrix}
    Sem_1,Sem_2,...,Sem_w
\end{Bmatrix}$, retrieve the semantic signpost frames in the candidate semantic element matching truth that satisfies the following formula with the semantic information of the current frame $Sem 'j = (n '_j,x '_j,y'_j,w'_j,h'_j), j \in (1,2,...,w)$ as the final candidate semantic signpost matching frames $L =\begin{Bmatrix}
    l_{b_1},l_{b_2 },...,l_{b_k}
\end{Bmatrix}$:
\begin{equation}
    \xi >\frac{\displaystyle\sum_{j}^{w}\sqrt{(x'_j-x_j)^2+(y'_j-y_j)^2+(w'_j-w_j)^2+(h'_j-h_j)^2}}{w}
\end{equation}
where $\xi$ is the set pixel frame deviation threshold, then, retrieve the semantic element frame in $L$ that is closest to the current frame as the semantic signpost matching frame $l_s$ that matches successfully, and obtain the semantic signpost matching frame bit posture $\mathbf{T}_w^{c_s}$ and geographic coordinates $\mathbf{E} = (x_s,y_s,z_s)^\mathrm{T}$. If the cumulative error of the camera position deduced by the visual odometer is large at this time, the cumulative error can be significantly corrected by directly assigning the semantic roadmap matching frame position to the current frame and initializing the visual odometer with the current frame as the first frame. If the cumulative error of camera position is small, the key frame position and map point coordinates are micro-corrected by our designed residual model.
\subsection{Multi-Sensor Fusion and Mapping}
Let the $i$-th map point be $\mathbf{P}_w^i = (x_w^i,y_w^i,z_w^i)^\mathrm{T}$, and the coordinate of the matched pixel be ${z}_x^i = (u^i,v^i)$, The camera pose for the current frame is $\mathbf{T}_w^{c_x} = [\mathbf{R}_w^{c_x}, \mathbf{t}_w^{c_x}] $, by the following formula:
\begin{equation}
    \mathbf{P}_c^i=\mathbf{R}_w^{c_x}\mathbf{P}_w^i+\mathbf{t}_w^i
\end{equation}
\begin{equation}
    z_c^i \begin{pmatrix}
u^i
\\ v^i
\\ 1
\end{pmatrix}
=\mathbf{K}\mathbf{P}_c^i
\end{equation}

$\mathbf{P}_c^i$ is projected to the pixel coordinate system with the camera internal reference matrix $\mathbf{K}$ according to the principle of small hole imaging. The Pose and Sem vectors for each frame are optimized by projecting the vectors onto the corresponding co-visual frames through the inverse projection residual model for each element. Constrain the current frame pose using semantic element matching frame poses to construct the observation equations:
\begin{equation}
    \mathbf{t}_w^{c_s} = \mathbf{t}_w^{c_x}+v_s
\end{equation}
where $v_s$ belongs to the noise and $ v_s\sim N(0,\varSigma_{sem})$, $\varSigma_{sem}$ is the covariance matrix.
Based on the classical beam leveling method, we introduce a new residual optimization term to optimize the camera position and map point coordinates, and construct the objective function as follows:
\begin{equation}
\begin{aligned}
  &\begin{Bmatrix}
\mathbf{p}_i,\mathbf{R}_x,\mathbf{t}_x|i \in P_L,x \in K_L
\end{Bmatrix}=&\\&\mathop{\arg\min}\limits_{\mathbf{p}_i,\mathbf{R}_x,\mathbf{t}_x}{(\displaystyle\sum_{x\in K_L}^{}\displaystyle\sum_{i \in P_L}^{}\rho(\parallel E_p(x,i) \parallel ^2_ \varSigma)+\displaystyle\sum_{k \in K_s}^{}\displaystyle\sum_{s\in T_s}^{}\rho_s(\parallel E_s(s,k)\parallel^2_{\varSigma_{sem}} ) }
\end{aligned}
\end{equation}
\begin{equation}
    \begin{aligned}
        &\parallel E_p(x,i)\parallel ^2_\varSigma =\parallel z_x^i-F_s(\mathbf{R}_w^{c_x}\mathbf{P}_w^i+\mathbf{t}_w^{c_x}
)\parallel ^2_\varSigma \\
=&(z_x^i-F_s(\mathbf{R}_w^{c_x}\mathbf{P}_w^i+\mathbf{t}_w^{c_x}))^\mathrm{T}\varSigma^{-1}(z_x^i-F_s(\mathbf{R}_w^{c_x}\mathbf{P}_w^i+\mathbf{t}_w^{c_x}))
    \end{aligned}
\end{equation}

\begin{equation}
    \begin{aligned}
        \parallel E_s(s,k)\parallel ^2_{\varSigma_{sem}} =\parallel \mathbf{t}_w^{c_s}-\mathbf{t}_w^{c_k}\parallel ^2_{\varSigma_{sem}}=(\mathbf{t}_w^{c_s}-\mathbf{t}_w^{c_k})^\mathrm{T}(\varSigma_{sem})^{-1}(\mathbf{t}_w^{c_s}-\mathbf{t}_w^{c_k})
    \end{aligned}
\end{equation}
where $ E_p(x ,i)$ is the error term constructed based on the reprojection error, $E_s(s ,k)$ is the error term constructed based on the semantic signpost matching frames, $F_s$ is the constraint function, $\rho$, $\rho_s$ are the robust kernel functions used to weaken the inferior error edges affecting the optimization results and make the optimization algorithm more robust, $P_L$ is the set of map points, $K_L$ is the set of local keyframes, $K_s$ are keyframes with corresponding semantic signpost matching frames, $T_s$ is the semantic element frame set, there are observation constraints between map points and keyframes, and there are observation constraints between $K_s$ and $T_s$. $\varSigma$, $\varSigma_{sem}$ are the covariance matrices of the reprojection error term and the covariance matrices of the semantic element frame error term for residual modeling, respectively, which represent the estimation of the accuracy of each constraint.
\subsection{Joint Optimization}
According to the exact camera pose $T_w^{c_x}$, the coordinate $O_w$ of the current vehicle in the world coordinate system is obtained, and the $n$ semantic landmark frames closest to the vehicle position are searched in the semantic element reference database, and the set $P = \begin{Bmatrix}
    p_1,p_2,... ,p_n
\end{Bmatrix}$ is constructed according to their coordinates in the world coordinate system, and the associated geographic coordinates construct the set $D = \begin{Bmatrix}
    d_1,d_2,... ,d_n
\end{Bmatrix}$, and finally construct the least squares problem to solve the transformation matrix from the world coordinate system to the geographic coordinate system:
\begin{equation}
    \min_{\mathbf{R}_w^g,\mathbf{t}_w^g}\frac{1}{2}\displaystyle\sum_{i=1}^{n}\parallel \mathbf{d}_i-(\mathbf{R}_w^g\mathbf{p}_i+\mathbf{t}^g_w)  \parallel ^2
\end{equation}
using the obtained $(\mathbf{R}_w^g,\mathbf{t}_w^g)$ to transfer $O_w$ to the geographic coordinate system, the position coordinates of the vehicle in the geographic coordinate system are output, thus realizing the high-precision navigation and positioning of the vehicle.\begin{table*}[!ht]
    \centering
    \caption{EVALUATION RESULTS OF THE ELEMENTS OF THE HIGHWAY.}

    \begin{tabular}{cccc|ccc|ccc}
    \toprule
        \multirow{2}{*}{Method} ~ &  \multirow{2}{*}{Semantic element} ~& \multicolumn{2}{c}{Existence}  ~& \multicolumn{3}{c}{MAE (m)} ~& \multicolumn{3}{c}{MRE (m)} \\ 
        ~ &  & Recall & Precision & Lateral & Longitudinal & Altitudinal & Lateral & Longitudinal & Altitudinal  \\ \midrule
        \multirow{4}{*}{DT-Loc (Camera) \cite{r30}}  & Lane Boundary & 90.53 & 94.45 & 1.52 & 1.59 & -1.77 & -0.53 & 0.43 & -0.32  \\ 
        ~ & Arrow & 84.58 & 88.19 & -1.21 & 3.60 & -2.32 & -0.28 & 0.36 & 0.40  \\ 
        ~ & Sign & 67.14 & 78.53 & -1.33 & -5.11 & -3.52  & -0.33 & -0.75& 0.45 \\ 
        ~ & Roadside Barrier & 83.57 & 87.62 & 1.09 & 2.11 & -2.37 & 0.52 & -0.67& 0.49  \\ 
        \midrule
        \multirow{4}{*}{ORB-SLAM3 (Camera) \cite{r5}}  & Lane Boundary & 91.89 & 96.47 & 1.15 & 1.20 & -1.92 & -0.14 & 0.38& -0.04  \\ 
        ~ & Arrow & 89.65 & 92.73 & -1.08 & 3.56 & -2.18 & -0.15 & 0.29& 0.35  \\ 
        ~ & Sign & 76.98 & 78.25 & -1.06 & -4.49 & -2.60  & -0.27 & -0.53& 0.47  \\ 
        ~ & Roadside Barrier & 81.61 & 87.23 & 1.13 & 1.74 & -1.97 & -0.02 & -0.48& 0.13  \\ 
        \midrule
        \multirow{4}{*}{\textbf{Ours}} & Lane Boundary & 95.88 & 97.24 & -1.02 & 1.15 & -0.67 & 0.03 & 0.09& 0.02  \\ 
        ~ & Arrow & 91.45 & 96.98 & -1.07 & 1.20  & -0.69 & 0.03 & 0.07& 0.02  \\ 
        ~ & Sign & 75.87 & 81.96 & -1.04 & 1.49 & -0.80  & 0.08 & -0.05& 0.06  \\ 
        ~ & Roadside barrier & 85.45 & 89.33 & -1.04 & 1.31 & -0.59 & 0.01 &-0.03&  0.05  \\ \bottomrule
    \end{tabular}
    \label{table1}
\end{table*}

\section{Experiments}
\begin{figure}[h!]
	\begin{minipage}{0.49\linewidth}
		\vspace{3pt}
		\centerline{\includegraphics[width=\textwidth]{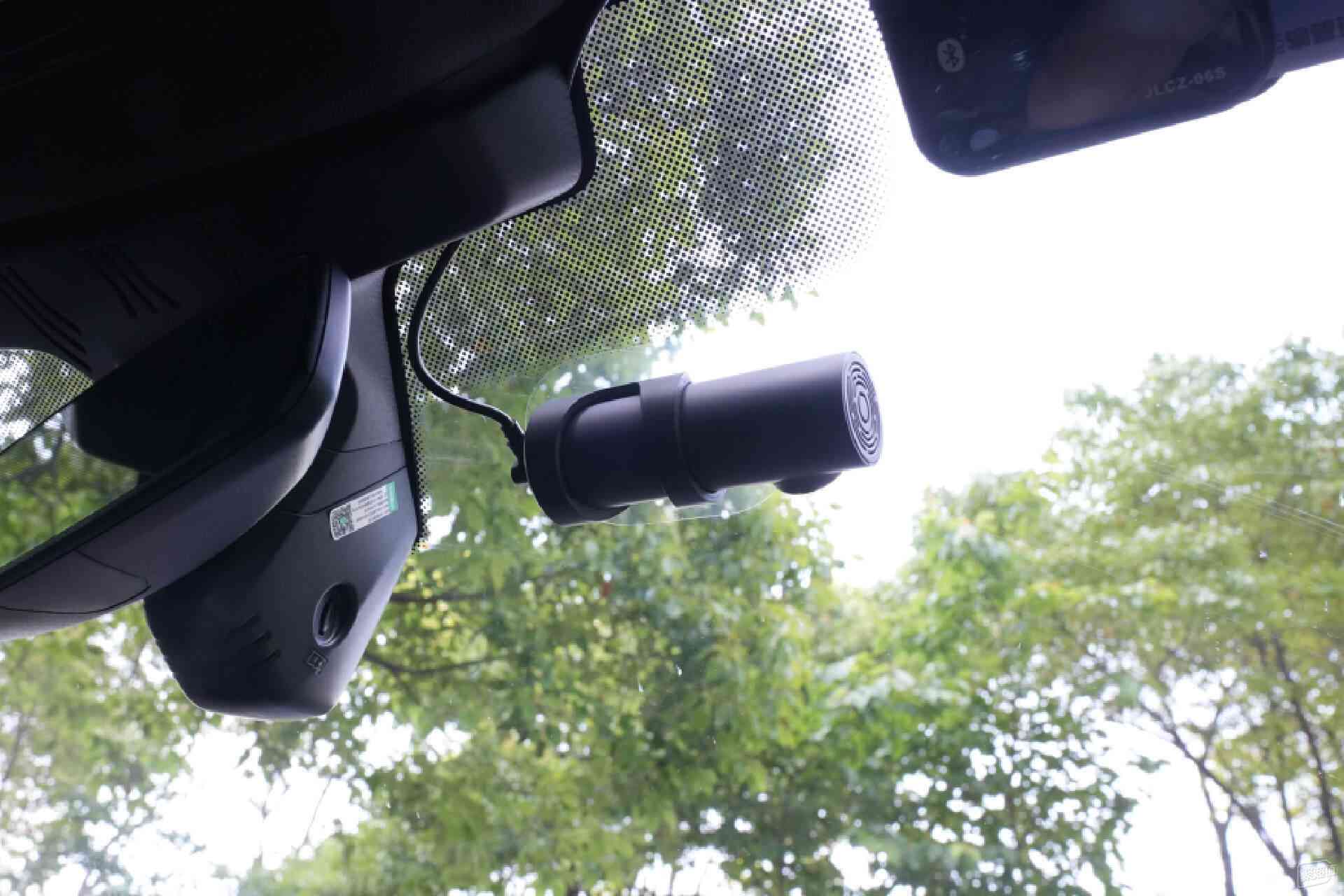}}

	\end{minipage}
	\begin{minipage}{0.48\linewidth}
		\vspace{3pt}
		\centerline{\includegraphics[width=\textwidth]{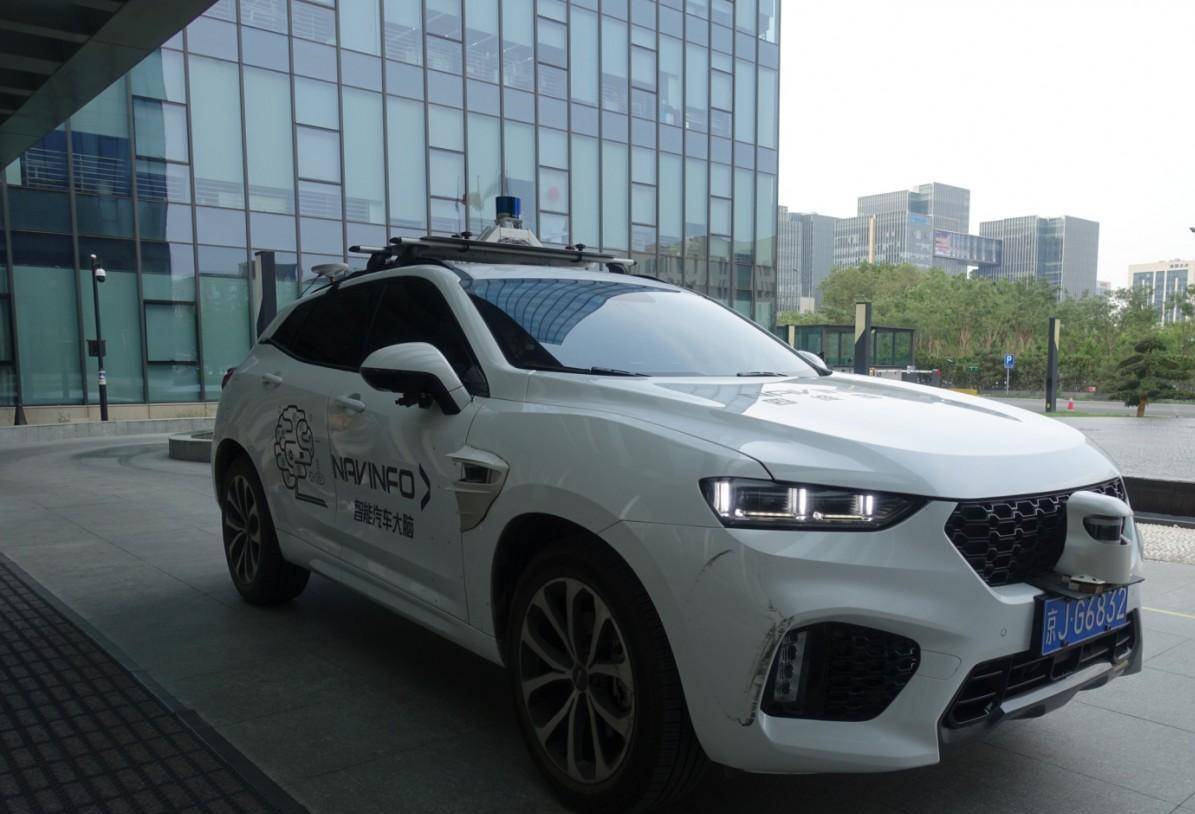}}
	\end{minipage}
	\caption{The test platform used in our experiments.}
	\label{cs}

\end{figure}

Our test vehicle platform consists of a rolling-shutter camera supporting video recording at resolutions up to 3840×2160, at which the video frame rate is 30 frames per second, an IMU, and a wheel encoder running at 100Hz and 50Hz, respectively. The vehicle is equipped with a NovAtel PwrPak7D-E1 INS module to obtain pseudo-ground-truth trajectories, achieving an accuracy of less than 10 cm when RTK is enabled. Fig. \ref{cs} shows our experimental equipment and test vehicles. We didn't use lidar in our experiment. The program is deployed on the Rv1126 chip, it is based on the quad-core ARM Cortex-A7 core, built-in 2T computing power NPU, and supports 4K30FPS H.264/H.265 video codec.

\subsection{Quantitative Analysis}
We tested our approach in different scenarios in Shanghai, China. The first scenario is on a normal road in the city, with a total length of 31.2 km, and contains multilane and intersection scenarios. The second case is a round trip on the highway, which is 168.5 km one way and 335 km in total. The variables in \cite{r5} and \cite{r30} are used in the experiments, using only the monocular camera method, to adapt to our test environment and data, and to use it as a baseline for a fair comparison. We report Precision, Recall, Mean Relative Error (MRE), and Mean Absolute Error (MAE) on partial map semantic elements, where the MAE is calculated as the difference between the distance between the position of the reported element and the actual position of the map element, including lateral, longitudinal, and altitudinal errors. Plus or minus represents the direction.\begin{figure*}[h!]
	\begin{minipage}{0.32\linewidth}
		\vspace{3pt}
		\centerline{\includegraphics[width=\textwidth]{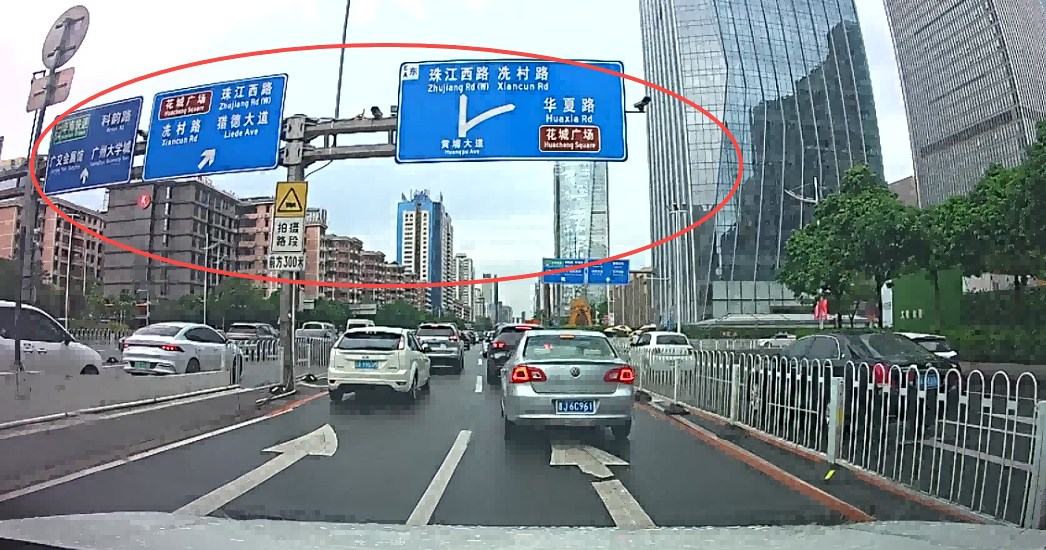}}
		\centerline{(a) Normal Road Scenario \#1}
	\end{minipage}
	\begin{minipage}{0.34\linewidth}
		\vspace{3pt}
		\centerline{\includegraphics[width=\textwidth]{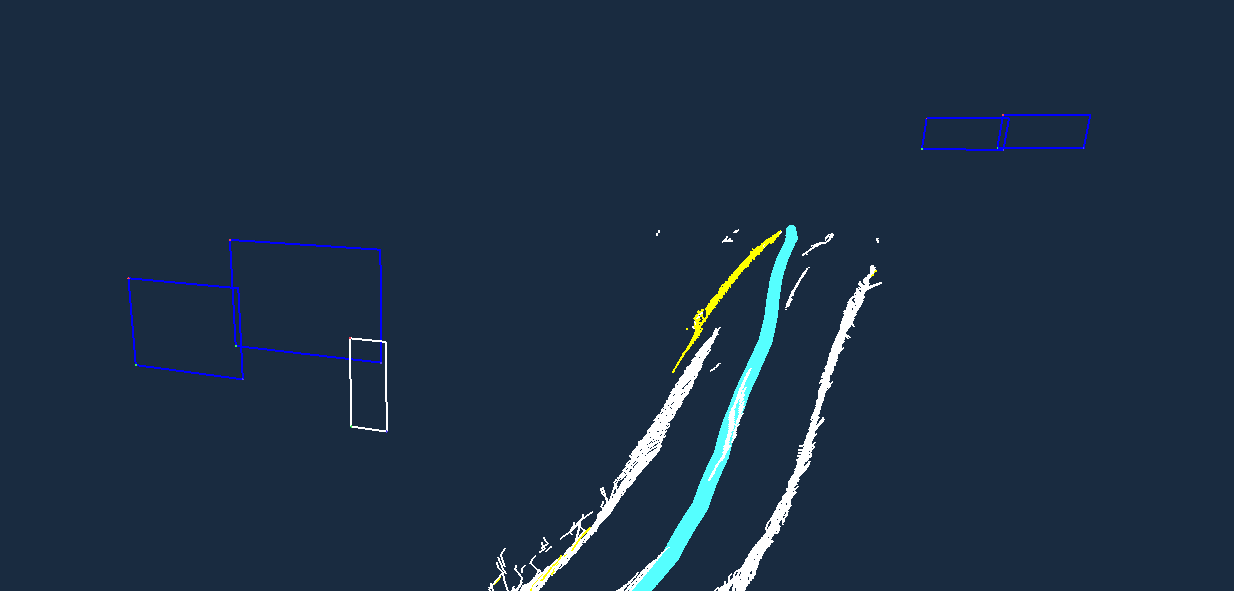}}
		\centerline{(b) Before Optimization}
	\end{minipage}
	\begin{minipage}{0.33\linewidth}
		\vspace{3pt}
		\centerline{\includegraphics[width=\textwidth]{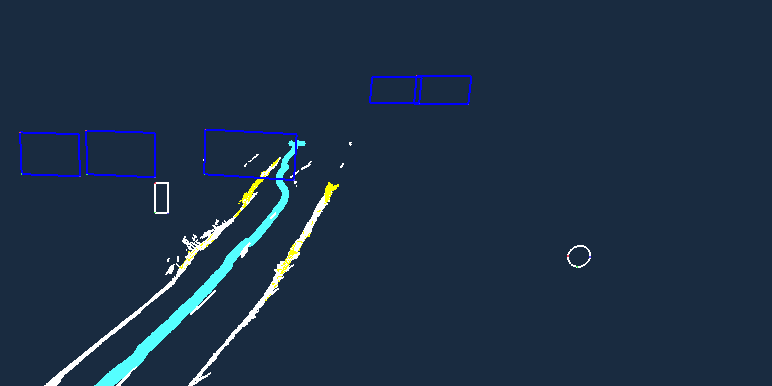}}
		\centerline{(c) After Optimization}
	\end{minipage}
 
	\caption{Example 1: Urban canyon, GPS signal is poor, the relative position error of traffic signs is poor before optimization, and there are missed recalls due to accuracy problems, and the relative position error is significantly reduced after optimization.}
	\label{fig2}
\end{figure*} 

\begin{figure*}[h!]
	\begin{minipage}{0.43\linewidth}
		\vspace{3pt}
		\centerline{\includegraphics[width=\textwidth]{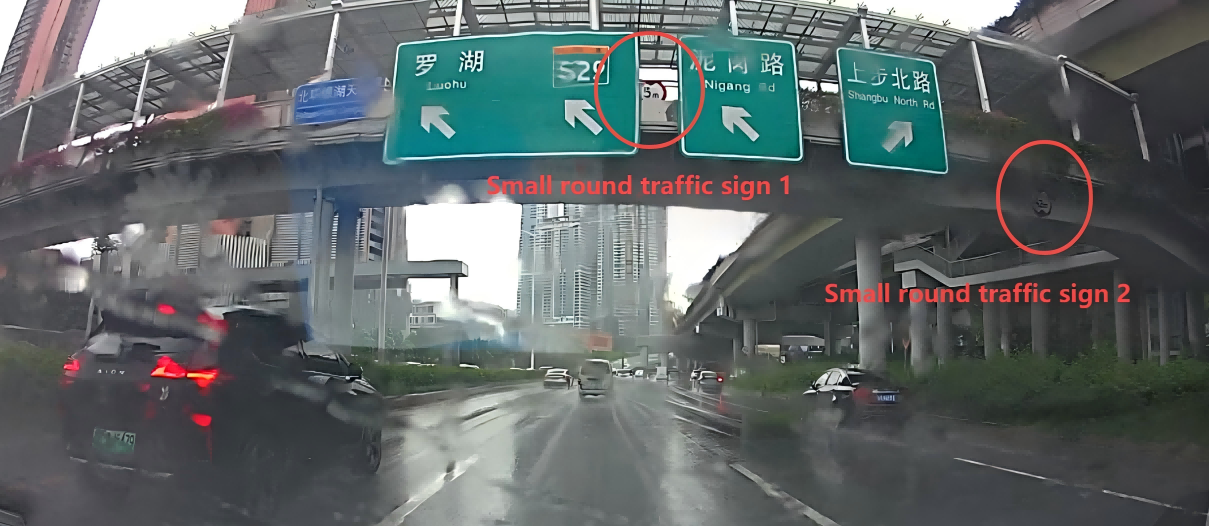}}
		\centerline{(a) Normal Road Scenario \#2}
	\end{minipage}
	\begin{minipage}{0.27\linewidth}
		\vspace{3pt}
		\centerline{\includegraphics[width=\textwidth]{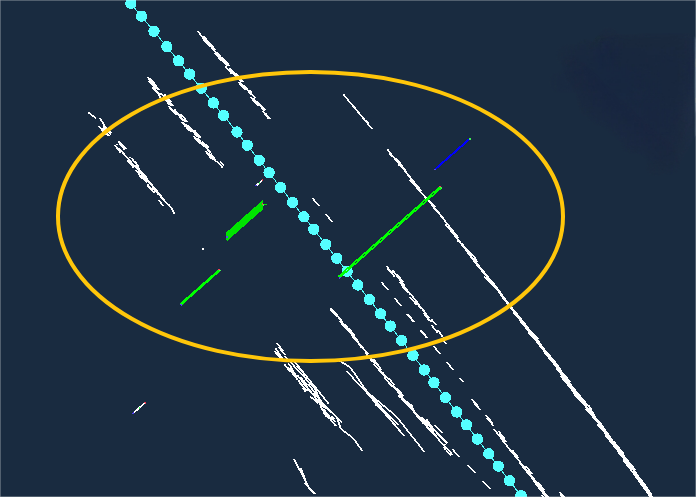}}
		\centerline{(b) Before Optimization}
	\end{minipage}
	\begin{minipage}{0.28\linewidth}
		\vspace{3pt}
		\centerline{\includegraphics[width=\textwidth]{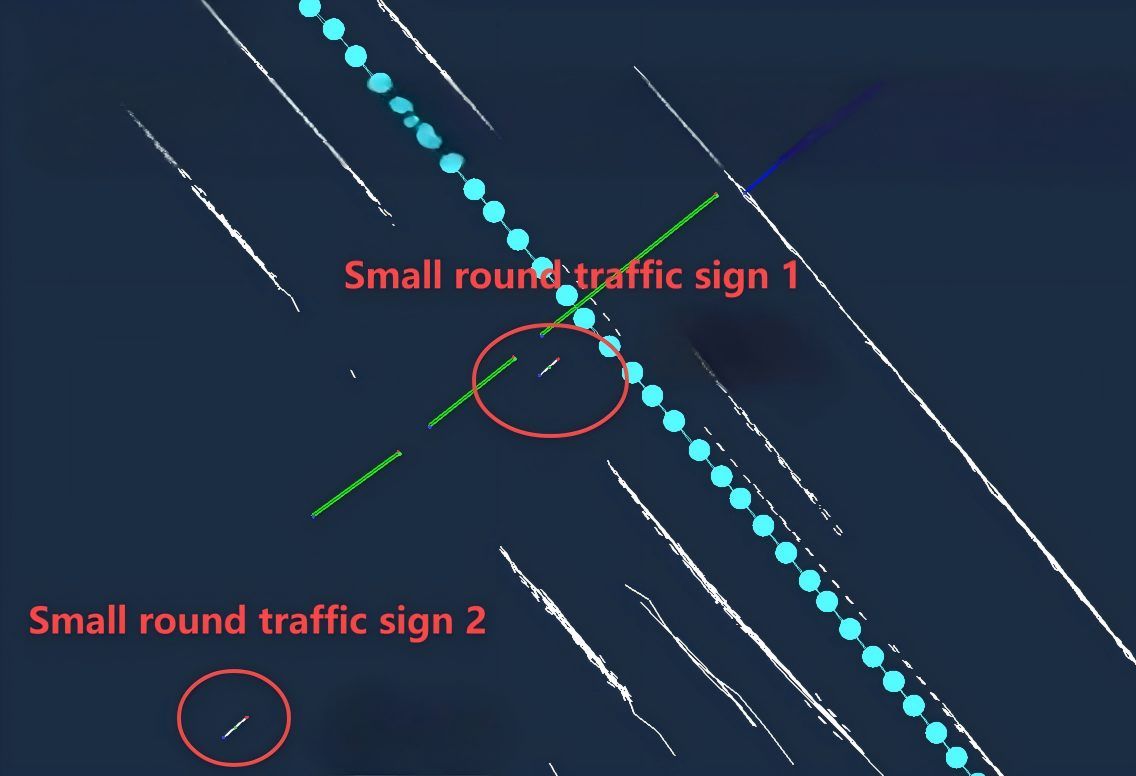}}
		\centerline{(c) After Optimization}
	\end{minipage}
 
	\caption{Example 2: Normal roads in the city are occluded, and the longitudinal relative position error between multiple traffic signs is large before optimization. After optimization, not only the relative relationship between large traffic signs is good, but also the relative position error between small traffic signs is low.}
    \vspace{-2mm}
	\label{fig3}
\end{figure*}

\begin{figure*}[h!]
\centering
	\begin{minipage}{0.41\textwidth}
		\vspace{3pt}
		\centerline{\includegraphics[width=\textwidth]{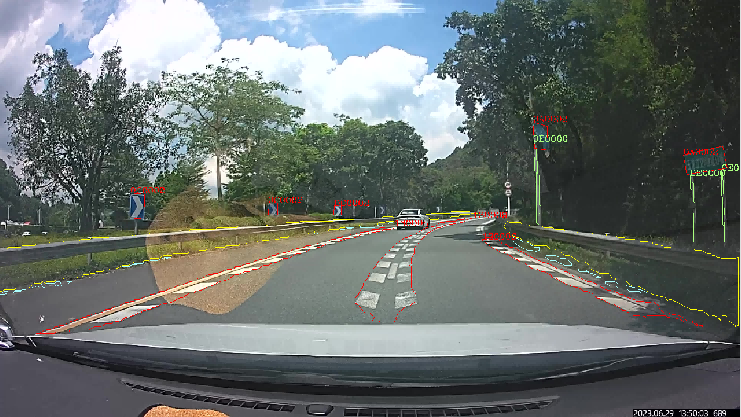}}
		\centerline{(a) Highways Scenario \#1}
	\end{minipage}
	\begin{minipage}{0.57\linewidth}
		\vspace{3pt}
		\centerline{\includegraphics[width=\textwidth]{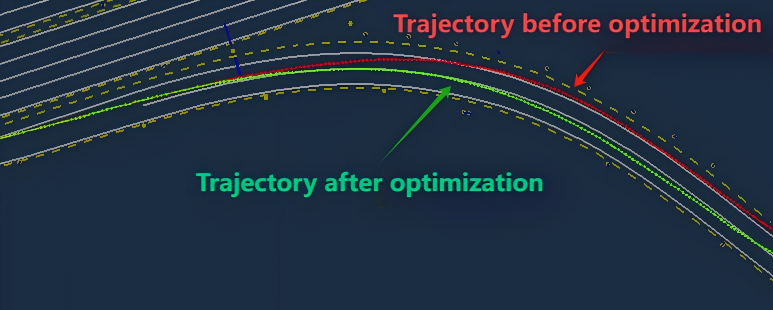}}
		\centerline{(b) Trajectory Comparison}
	\end{minipage}
	\caption{Example 3: Highways curve scenario, poor combined navigation accuracy, trajectory deviation to outside the lane, trajectory before optimization is shown in red, trajectory after optimization is shown in green.}
	\label{fig4}
\end{figure*}

\begin{figure*}[h!]
\centering
	\begin{minipage}{0.49\linewidth}
		\vspace{3pt}
		\centerline{\includegraphics[width=\textwidth]{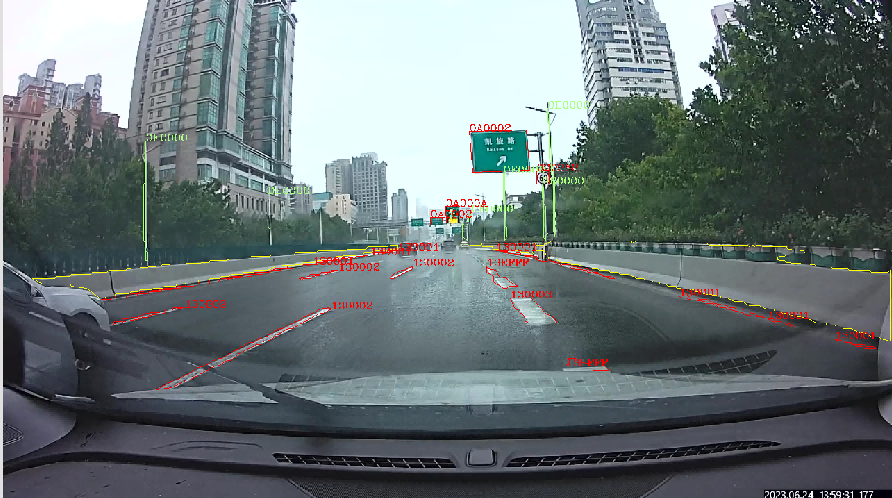}}
		\centerline{(a) Highways Scenario \#2}
	\end{minipage}
	\begin{minipage}{0.48\linewidth}
		\vspace{3pt}
		\centerline{\includegraphics[width=\textwidth]{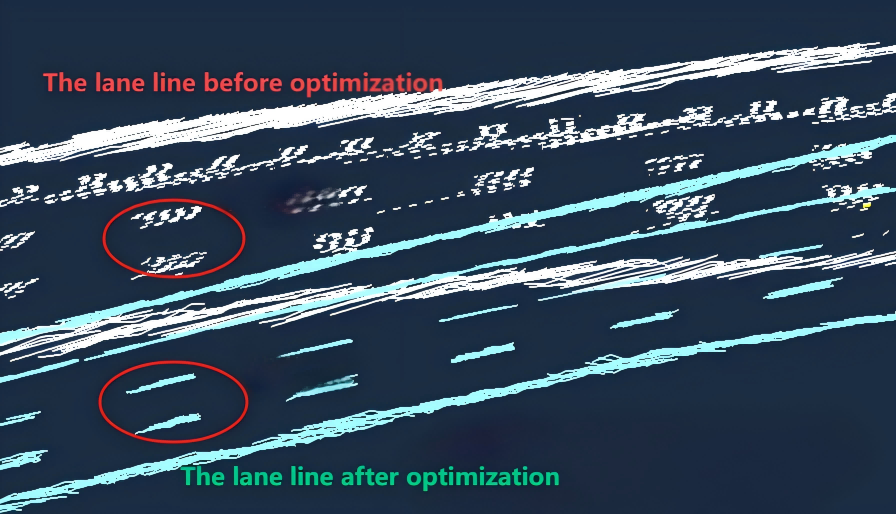}}
		\centerline{(b) Trajectory Comparison}
	\end{minipage}
	\caption{Example 4: Highway scenario, the attitude before optimization is abnormal, and the lane lines diverge significantly. The optimized lane lines are clustered clearly and with higher accuracy. The optimized front lane line is shown in white and the optimized back lane line is shown in blue.}
	\label{fig5}
    \vspace{-2mm}
\end{figure*}

The MRE is calculated as the difference between the distance between two objects and the corresponding distance between two objects on the map, including the lateral, longitudinal, and altitudinal errors. The negative values for the lateral, longitudinal, and altitudinal correspond to the left, rear, and below the reference line, respectively. Table \ref{table1} shows the results of the quantitative evaluation of the results. It can be seen that our method is better than the baseline or achieves a comparable level in almost all four semantic elements, where the recall and precision of Lane Boundary are 5.35\% and 2.79\% higher than that of HDMapNet respectively. It is 3.99\% and 0.77\% higher than ORB-SLAM3, respectively. It should be noted that the relative error of all elements has a significant improvement over the baseline, with the overall MRE within 10 cm, while the MAE range is around 1 m. This indicates that our method has high positioning accuracy and reliability in complex environments, and it is verified that the ability to enhance positioning and recognition through continuous semantic information on the map makes the vehicle more accurate to navigate and locate in changing traffic scenes.

\subsection{Qualitative Analysis}
From a qualitative point of view, the recognition accuracy and recall rate of the methods in the baseline is relatively low, which may be because the sign of the traffic road is complex and diverse, and the common feature point method is easily affected by illumination and environment, resulting in reduced accuracy. Our system still maintains strong environment awareness in dynamic road environments, which is due to image perception combined with semantic localization optimization. Even in scenes with complex traffic or large illumination changes, the system can rely on the associated semantic information in the environment and map construction to improve the identification accuracy of road elements through optimization. This continuous construction based on a semantic map enables the system to capture better the details of the road and environment in real-time operation, provide more accurate positioning and path planning information, and improve the safety and reliability of the autonomous driving system.

\subsection{Ablation Study}
We have conducted a large number of experimental cases of ablation in real-world scenarios. In the urban road test sequence, Fig. \ref{fig2}(a) is one of the cases in the test sequence on urban roads, and sub-figure (b) is the measurement rate before optimization, which uses only combined navigation for localization and mapping, and it can be seen that the relative position error of the traffic signs is poor, and due to the precision problem, there are missed recall of distant traffic signs, and the trajectory changes are not accurate enough. After optimization, as shown in sub-figure (c), the longitudinal relative position error between traffic signs is significantly improved all of them are recalled, and the trajectory changes are more accurate, which shows that our proposed method can effectively reduce the relative position error between map elements. Fig. \ref{fig3} (a) shows another example of a normal road, where the combined navigation accuracy is poor when there is an occlusion over an urban canyon, the longitudinal relative position error of traffic signs is poor, and some traffic signs are not recognized, as shown in subfigure (b). After optimization, as shown in subfigure (c), the relative positions of all traffic signs are significantly improved with no missed recalls.

Fig. \ref{fig4} shows an example of a high-speed curve scene, the pre-optimization trajectory is shown in red, the camera trajectory deviates outside the lane due to the poor accuracy of the combined navigation, and the post-optimization trajectory can be kept within the current lane. The white trajectory on the right of Fig. \ref{fig5} shows that the lane lines are dispersed due to the abnormal camera position before optimization, while the lane lines are clustered after optimization with higher accuracy. 

These ablation studies provide strong evidence that the proposed method improves localization accuracy, enhances the robustness of map construction under noisy conditions, and guarantees more reliable performance even in fast-moving or complex scenes.
\section{Conclusion}
This paper presents a semantic SLAM algorithm for outdoor environments using a low-cost, low-power rolling shutter camera and a low-precision inertial navigation system. Our approach successfully integrates semantic road element detection with rolling-shutter compensation and low-precision INS data, demonstrating significant improvements in both accuracy and reliability, with enhanced recall (up to 5.35\%) and precision (up to 2.79\%) in semantic element detection while maintaining mean relative error within 10cm. The system shows particular robustness in challenging scenarios such as GPS-denied environments and complex urban settings, validating its practical viability for autonomous navigation applications. While opportunities remain for extending the system to handle more dynamic environments and improving real-time performance on embedded platforms, the demonstrated success using affordable hardware components marks a significant step toward making robust localization more accessible for widespread deployment. By bridging the gap between consumer-grade sensors and production-level performance requirements, our work provides a practical solution for large-scale applications in autonomous navigation, paving the way for future developments in cost-effective and reliable autonomous systems.


\section*{Acknowledgments}
This work was sponsored by Beijing Nova Program (No. 20240484616).

\balance
\bibliographystyle{IEEEtran}
\bibliography{IEEEexample}
\end{document}